\documentclass{article}
\usepackage[utf8]{inputenc} 
\usepackage[T1]{fontenc}    
\usepackage{hyperref}       
\usepackage{url}            
\usepackage{booktabs}       
\usepackage{amsfonts}       
\usepackage{nicefrac}       
\usepackage{microtype}      
\usepackage{amsfonts}
\usepackage{amssymb}
\usepackage{diagbox}
\usepackage{pstricks}
\usepackage{setspace}
\usepackage{amsthm}
\usepackage{subfigure}
\usepackage{graphicx}
\definecolor{content}{rgb}{0,0,1}
\definecolor{delete}{rgb}{1,0,0}
\definecolor{notsure}{rgb}{1,0,1}

\title{Distribution-dependent concentration inequalities for tighter generalization bounds}
\author{
Xinxing Wu\\
Department of Computer\\
Shanghai Technical Institute of\\
Electronics \& Information\\
Shanghai, 201411\\
\texttt{xinxingwu@yeah.net} \\
\and
Junping Zhang\thanks{Telephone: 86-21-55664503, Fax: 86-21-65654253.} \\
Shanghai Key Laboratory\\
of Intelligent Information Processing\\
School of Computer Science\\
Fudan University\\
Shanghai, 200433\\
\texttt{jpzhang@fudan.edu.cn} \\
}

\begin{document}

\maketitle

\begin{abstract}
Concentration inequalities are indispensable tools for studying the generalization capacity of learning models. Hoeffding's and McDiarmid's inequalities are commonly used, giving bounds independent of the data distribution. Although this makes them widely applicable, a drawback is that the bounds can be too loose in some specific cases.  Although efforts have been devoted to improving the bounds, we find that the bounds can be further tightened in some distribution-dependent scenarios and conditions for the inequalities can be relaxed. In particular, we propose four types of conditions for probabilistic boundedness and bounded differences, and derive several distribution-dependent extensions of Hoeffding's and McDiarmid's inequalities. These extensions provide bounds for functions not satisfying the conditions of the existing inequalities, and in some special cases, tighter bounds. Furthermore, we obtain generalization bounds for unbounded and hierarchy-bounded loss functions. Finally we discuss the potential applications of our extensions to learning theory.
\end{abstract}

\section{Introduction}\label{sec1}
Concentration inequalities play a crucial role in statistical learning theory because they are useful for deriving the generalization capacity of learning models. Generally,  they can be used to estimate the deviations between empirical risk and expectation risk \cite{Vapnik}. Some important learning theories such as Rademacher complexity have been developed by applying concentration inequalities to bound such deviations \cite{Koltchinskii,Bartlett,Mendelson,Bousquet,Schapire,Freund,Boucheron2005Theory,Shalizi2013Predictive,Mohri2010Rademacher}.

Two commonly used concentration inequalities in learning theory are Hoeffding's \cite{Hoeffding} and McDiarmid's inequalities \cite{McDiarmid}. Besides being used to analyze the algorithmic stability \cite{Bousquet,Kutin,Kutins,KontorovichA}, Hoeffding's and McDiarmid's inequalities, both giving bounds independent of the distribution, are powerful tools for estimating VC dimension and Rademacher complexity \cite{Yi,Lei2014Refined,Cao2012Generalization,Ying2010Rademacher,Gao2014Dropout}.

These two inequalities however have two major limitations: 1) they cannot deal with unbounded functions; 2) their bounds are weak for functions with a larger constant on a small exceptional set. If we generalize these inequalities to any distribution, the estimation of the deviations is likely to be loose. In the latter case, the bounds given by them will become less tight because the large constant dominates this bound. To address these issues, \cite{Kutins,Niyogi} proved two extensions of McDiarmid's inequality for strongly and weakly difference-bounded functions and used them to study the generalization capacity. \cite{KontorovichA} proved an extension of McDiarmid's inequality with the subgaussian diameter. Recently, \cite{Combes} proposed an extension of McDiarmid's inequality for functions with bounded differences on a high probability set and no restriction outside this set.  \cite{Warnke} extended McDiarmid's inequality by relaxing the Lipschitz condition since the approach only needs Lipschitz-bounds for changing one variable.

However, both the strong and weak bounded difference conditions proposed by \cite{Kutins,Niyogi} have their shortcomings in practice. The approach proposed by \cite{KontorovichA} requires an extra metric on the sample space and a bounded subgaussian diameter. The weaker Lipschitz condition given by \cite{Warnke} is only useful for bounded functions. Meanwhile, the bound discussed by \cite{Combes} can be further tightened, which will be detailed in Section~2. After exploring the assumptions of Hoeffding's and McDiarmid's inequalities, we propose some extensions to these two inequalities to treat the cases of probabilistic boundedness and bounded differences. Our results improve the bound in \cite{Combes} and the bounds in the original inequalities, and can also handle unbounded functions without introducing extra metrics.

{\bf Main results.} we prove some distribution dependent extensions of Hoeffding's and McDiarmid's inequalities and obtain tighter generalization bounds. In Theorem 1 and Corollary 1, we obtain some new inequalities for the cases of probabilistic boundedness and bounded differences. These extensions are distribution dependent, and consequently yield better estimation (For example, Theorem 2 and Corollary 2) for some examples in learning theory. 

{\bf Motivation.} The unbounded functions often occur in the analysis of regression and classification \cite{Marco2005Consistency,Cortes2013Relative,Dud2005Correcting,PeterM}. Since Hoeffding's and McDiarmid's inequalities provide bounds independent of the distribution, we expect that our proposed distribution-dependent bounds will be tighter for each specific case. 

{\bf Outline.} In Section~2, we analyze the disadvantages of Hoeffding's and McDiarmid's inequalities, and discuss related work. In Section~3, we introduce the basic notations and definitions. In Section~4, we show the limitations of the conditions in existing inequalities using two examples. Furthermore, we propose four assumptions about the probabilistic boundedness and bounded differences, and compare these differences with the previous bounded ones. In Section~5, we present the probabilistic extensions of Hoeffding's and McDiarmid's inequalities. In Section~6, we discuss the potential applications of our results in learning theory. We show how to use our results to analyze the generalization of learning models. We conclude in Section~7.

\section{Related Work}
Although Hoeffding's and McDiarmid's inequalities have achieved great success in learning theory,\cite{Kutin,Kutins,KontorovichA,Combes} have noted their limitations in applications due to the fact that these inequalities are distribution independent and cannot provide generalization bounds for unbounded loss functions. 

To address these issues, researchers have studied more general conditions under which concentration inequalities exist. Specifically, assuming that the function is bounded on one side, \cite{Bentkus2008An} gave an extension of Hoeffding's inequality to unbounded random variables with bounded mathematical expectation. \cite{Kutins,Niyogi} proved two extensions of McDiarmid's inequality to strongly and weakly difference-bounded functions (See Definitions 2 and 3 in Section~3) for the study of the generalization error. 

\cite{Kutins,Niyogi} assumed that there exist some constant vectors, e.g., $b$ and $c$, with $b_i\geq c_i$ for all $i=1,2,\ldots, n$, such that the function $f$ has $b$ bounded differences on a subset set $\mathcal{D}$ of $\mathcal{X}$ and $c$ bounded differences on the complement of the set $\mathcal{D}$. \cite{KontorovichA} noted that the strong and weak bounded difference conditions proposed by \cite{Kutins,Niyogi} have their limitations in practice and the bounds of the inequalities are uninformative if $b$ is infinite. In order to relax the difference-bounded conditions, \cite{KontorovichA} introduced the subgaussian diameter and proved an extension of McDiarmid's inequality using the subgaussian diameter. Nevertheless, the approach \cite{KontorovichA}  proposed requires an extra metric on the sample space and that the subgaussian diameter is bounded. Recently, \cite{Combes} developed more general difference-bounded conditions: the function $f$ has $c$ bounded differences on a high probability set $\mathcal{D}(\subset\mathcal{X})$ and is arbitrary outside of $\mathcal{D}$, the measure of which is controlled by a probability $p$  (This is similar to Assumption 3 in Section~4). Finally, \cite{Combes} proposed an extension of McDiarmid's inequality:{\small
\begin{equation}{\label{comb}}
{\rm{P}}(|f(X_{1},\ldots,X_{n})-{\rm{E}}(f(X_{1},\ldots,X_{n}|\mathcal{D}))|\geq t)\leq 2\cdot\Delta_{1}
\end{equation}}Here, $\Delta_{1}=p+{\rm{exp}}({-{2 ((t-p\cdot\overline{c})^{+})^2}/{(\sum_{i=1}^{n}c_{i}^{2}})})$, $\overline{c}=\sum_{i=1}^{n} c_{i}, (t-p\cdot\overline{c})^{+}=\max\{t-p\cdot\overline{c},0\}$. It is worth pointing out that the above bound in the equation (\ref{comb}) discussed by \cite{Combes} can be further tightened.

Different from the boundedness conditions previously discussed, our extensional conditions do not require 1) loss functions to be bounded as in \cite{Kutins}, and 2) the extra metrics as in \cite{KontorovichA}. Roughly speaking, they can be classified into two cases:

$\blacksquare$ Being similar to the boundedness conditions given by \cite{Combes} (See Assumptions 1 and 3 in Section~4). In this case, we will obtain a refined bound.

$\blacksquare$ Refining the boundedness and bounded differences (See Assumptions 2 and 4 in Section~4). In this case, we will obtain tighter bound. 

\section{Notations and Definitions}
In this section, we introduce notations and definitions.

$\blacksquare$ Let $I_{A}$ denote the indicator function.

$\blacksquare$ Let $\mathbb{N}$ be the set of natural numbers, $\mathbb{R}$ be the set of real numbers. Let $\mathbb{N}_{n}=\{1,2,\ldots,n\}, n\in\mathbb{N}$.

$\blacksquare$ Let $(\Omega,\mathcal{A},{\rm{P}})$ be a probability space, that is, $\Omega$ alone is called the sample space, $\mathcal{A}$ is a $\sigma$-algebra on $\Omega$, and ${\rm{P}}$ is a probability measure on $(\Omega, \mathcal{A})$. And $\Omega$ has the structure $\Omega=\mathcal{X}\times\mathcal{Y}$, where $\mathcal{X}$ and $\mathcal{Y}$ are the input space and output space respectively. The set $\emptyset$ denotes the empty set.

$\blacksquare$ Let $\mathcal{F}$ be the set of all measurable functions $f: \mathcal{X}\longrightarrow\mathcal{Y}$. Assume that $\mathcal{H}$ is a subset of $\mathcal{F}$, i.e., $\mathcal{H}\subset\mathcal{F}$, the set $\mathcal{H}$ is called the hypothesis class.

$\blacksquare$ Let $S=\{z_{i}=(x_i,y_i),i\in\mathbb{N}_{n}\}$ be a finite set of labeled training samples, and assume that these samples are independent and identically distributed (i.i.d.) according to $\rm{P}$. Denote the bold letter as a vector, for example, the bold ${\bf z}$ presents a vector $(z_{1},z_{2},\ldots,z_{n})$.

$\blacksquare$ Let $L$ be the loss function, $L: \mathcal{Y}\times\mathcal{Y}\longrightarrow[0,+\infty]$, and the loss of $f$ on a sample point $z=(x,y)$ is defined by $Q(f,z)=L(f(x),y)$. We can see that the function $Q$ is nonnegative, but not necessarily bounded.  Three well-known examples for this function often used in machine learning domain are the absolute loss $Q(f,z)=|f(x)-y|$, squared loss $Q(f,z)=(f(x)-y)^{2}$ and $\log$ loss $Q(f,z)=-\log {p_{f}(y|x)}$ \cite{KontorovichA,PeterM}. Here $(x,y)\in\mathcal{X}\times\mathcal{Y}$ and $\{p_{f}|f\in\mathcal{F}\}$ is a statistical model of conditional densities for $y|x$.

In learning theory, one of the goals is to find a function $h$ in hypothesis space $\mathcal{H}$ that minimizes the following generalization error ${\rm{E}}(Q)\triangleq \int_{\Omega}Q(h,z)d{\rm{P}}$. Generally speaking, the distribution $\rm{P}$ in the equation ${\rm{E}}(Q)$ is unknown. Rather than minimizing ${\rm{E}}(Q)$, we usually minimize the following training error:
\begin{equation}
{\rm{E}}_{n}(Q)\triangleq({1}/{n})\sum_{i=1}^{n}Q(h,z_{i})
\end{equation}

In this paper, we are interested in the uniform estimation of ${\rm{E}}(Q)-{\rm{E}}_{n}(Q)$.

{\bf{Definition 1 (Uniformly difference-bounded \cite{Kutins,Niyogi})}} Let $g:\prod_{k=1}^{n}\Omega_{k}\rightarrow\mathbb{R}$ be a function. We say that $g$ is uniformly difference-bounded by $\{c_{k},k\in\mathbb{N}_{n}\}$, if the following holds:

For any $k\in\mathbb{N}_{n}$, if $\omega,\omega'\in\prod_{k=1}^{n}\Omega_{k}$ differ only in the $k$th coordinate, that is, there exists $\omega_{1},\ldots,\omega_{n},\omega'_{k}\in\Omega$, s.t. $\omega=(\omega_{1},\ldots,\omega_{k},\ldots,\omega_{n})$ and $\omega'=(\omega_{1},\ldots,\omega'_{k},\ldots,\omega_{n})$, then we have $|g(\omega)-g(\omega')|\leq c_{k}$.

{\bf{Definition 2 (Strongly difference-bounded \cite{Kutins,Niyogi})}} Let $g:\prod_{k=1}^{n}\Omega_{k}\rightarrow\mathbb{R}$ be a function. We say that $g$ is strongly difference-bounded by $(\{b_{k},k\in\mathbb{N}_{n}\},\{c_{k},k\in\mathbb{N}_{n}\},\delta)$, if the following holds:

There exists a \lq\lq bad\rq\rq subset $B\subset\prod_{k=1}^{n}\Omega_{k}$, where $\delta={\rm{P}}(B)$. For any $k\in\mathbb{N}_{n}$, if $\omega,\omega'\in\prod_{k=1}^{n}\Omega_{k}$ differ only in the $k$th coordinate and $\omega\not\in B$, then $|g(\omega)-g(\omega')|\leq c_{k}$; if $\omega,\omega'\in\prod_{k=1}^{n}\Omega_{k}$ differ only in the $k$th coordinate, then $|g(\omega)-g(\omega')|\leq b_{k}$.

{\bf{Definition 3 (Weakly difference-bounded \cite{Kutins,Niyogi})}} Let $g:\prod_{k=1}^{n}\Omega_{k}\rightarrow\mathbb{R}$ be a function. We say that $g$ is weakly difference-bounded by $(\{b_{k},k\in\mathbb{N}_{n}\},\{c_{k},k\in\mathbb{N}_{n}\},\delta)$, if the following holds:

For any $j\in\mathbb{N}_{n}$, we have{\small
\begin{equation}{\label{equ7}}
{\rm{P}}((\omega,v)\in(\prod_{j=1}^{n}\Omega_{j})\times\Omega_{j}\arrowvert  \ \ \ |g(\omega)-g(\omega')|>c_{j})\leq\delta
\end{equation}}
where $\omega'\in\prod_{k=1}^{n}\Omega_{k}$, $\omega_{j}^{'}=v$ and $\omega_{i}^{'}=\omega_{i}$ for $i\not = j$.

For any $\omega$ and $\omega'$ differing only in the $k$th coordinate,  moreover, $|g(\omega)-g(\omega')|\leq b_{k}$.

{\bf{Note 1}} The equation (\ref{equ7}) means that if we construct $\omega'\in\prod_{k=1}^{n}\Omega_{k}$ by replacing the $k$th entry of $\omega\in\prod_{k=1}^{n}\Omega_{k}$ with $v$, then $|g(\omega)-g(\omega')|\leq c_{k}$ holds for all but a $\delta$ fraction of the choices.

For the discussion of later sections and to be self-contained,  we provide the original forms of Hoeffding's and McDiarmid's inequalities as follows:

{\bf Hoeffding's inequality \cite{Hoeffding}} Let $X_{1},X_{2},\ldots,X_{n}$ be independent random variables on a probability space $(\Omega,\mathcal{A},{\rm{P}})$, s.t. $X_{i}\in[a_{i},b_{i}], i=1,2,\ldots,n$. Set $S_{n}=\sum_{i=1}^{n}X_{i}$. Then, for all $t\geq 0$ we have
{\small\begin{equation}{\label{hoe}}
\displaystyle {\rm{P}}(|S_{n}-{\rm{E}}(S_{n})|\geq t)\leq 2\cdot\Delta_{2}
\end{equation}}where $\Delta_{2}={\rm{exp}}({{-2 t^2}/({\sum_{i=1}^{n}(a_{i}-b_{i})^{2}})})$.

{\bf McDiarmid's inequality \cite{McDiarmid}} Let $X_{1},X_{2},\ldots,X_{n}$ be independent random variables on a probability space $(\Omega,\mathcal{A},{\rm{P}})$. Then, for all $t\geq 0$ we have{\small
\begin{equation}{\label{mcd}}
\displaystyle{\rm{P}}(|f(X_{1},\ldots,X_{n})-{\rm{E}}(f(X_{1},\ldots,X_{n}))|\geq t)
\leq 2\cdot\Delta_{3}
\end{equation}}where $\Delta_{3}={\rm{exp}}\left({{-2 t^2}/\left({\sum_{i=1}^{n}c_{i}^{2}}\right)}\right)$, the function $f$ is a real-valued function of the sequence $X_{1},X_{2},\ldots,X_{n}$, s.t. $|f(x)-f(x')|\leq c_{i}$, whenever $x$ and $x'$ differ only in the $i$th coordinate, $i=1,2,\ldots,n$, the uniformly difference bounded function  $f$ is uniformly difference-bounded. 

\section{Addressing the Limitations of Previous Concentration Inequalities}
In this section, we analyze the conditions of the inequalities in previous works, and show that they have limitations in some examples. At the end of this section, we discuss several general assumptions for the boundedness conditions of concentration inequalities.

\subsection{Limitations in two cases}
Here, we analyze two examples to illustrate the limitations of previous concentration inequalities.

{\bf{Example 1}}  Let $\omega\in\mathbb{N}$, set $X(\omega)\triangleq \omega\cdot I_{(n_0,+\infty)}(\omega)$. For all $\omega=k,k\in[1,+\infty)$, if we set  ${\rm{P}}(\omega=k)\triangleq{6}/\left({\pi^2 k^2}\right)$, then we have the identity $\sum_{k=1}^{\infty}{\rm{P}}(\omega=k)=\sum_{k=1}^{\infty}\left({6}/{\pi^2 k^2}\right)=1$. By the above definition of $X(\omega)$, it follows that
${\rm{E}}(X\cdot I_{[1,n_0]})=\sum_{k=1}^{n_0}\left({0\cdot6}/{\pi^2 k^2}\right)=0$
and ${\rm{E}}(X\cdot I_{(n_0,\infty)})=\sum_{k=n_0}^{\infty}\left({k\cdot6}/{\pi^2 k^2}\right)=\left({6}/{\pi^2}\right)\sum_{k=n_0}^{\infty}\left({1}/{ k}\right)=\infty$.

{\bf{Note 2}} In this example, the random variable $X(\omega)$ is unbounded, and thus does not satisfy the condition of Hoeffding's inequality. If the random variable $X(\omega)$ is limited in a certain range (for example, $\omega\in[1,n_{0}]$), the condition of Hoeffding's inequality will be satisfied.

{\bf{Example 2}}  Let $\omega\in\mathbb{N}$, set $f(X_{1}(\omega),\ldots,X_{n}(\omega))\triangleq M^{n}I_{\omega=M}(\omega)$. Here, we set $X_{i}(\omega)=\omega, \omega\in\mathbb{N}_{M}$, $M$ is a constant, $i\in\mathbb{N}_{n}$.

For all $\omega=k,k\in\mathbb{N}_{M}$, we assume that ${\rm{P}}(\omega=k)\triangleq {1}/(M)$ where $M$ presents the cardinality of $\mathbb{N}_{M}$. By the above definition of $f(X_{1}(\omega),\ldots,X_{n}(\omega))$, it follows that ${\rm{E}}(f(X_{1},\ldots,X_{n})\cdot I_{[1,M-1]})=\sum_{k=1}^{M-1}\left({0\cdot 1}/{M}\right)=0$ and ${\rm{E}}(f(X_{1},\ldots,X_{n})\cdot I_{M})=\left({M^{n}\cdot 1}/{M^{n}}\right)=1$.

{\bf{Note 3}} In this example, we know that $f$ is not uniformly differences-bounded, failing to the condition of McDiarmid's inequality. But if the sample space is limited in a certain range (for example, $\omega\in[1,M-1]$), the condition of McDiarmid's inequality is satisfied. In addition, we observe that $f$ is neither weakly differences-bounded nor strongly differences-bounded.

\subsection{Assumptions}
The aforementioned examples 1 and 2 do not satisfy various existing definitions of the boundedness and bounded differences. A possible reason behind is that these existing definitions either are sort of restrictive or neglect the robustness of learning theory in some cases. Therefore, four assumptions below are proposed to alleviate these issues. 

{\bf{Assumption 1 ($p_{i}$ bounded)}} Let $X_{i}$ be the independent random variable on a probability $(\Omega_{i},\mathcal{A}_{i},{\rm{P}}_{i}), i\in\mathbb{N}_{n}$, s.t. ${\rm{P}}_{i}(a_{i}\leq X_{i}(\omega)\leq b_{i})=p_{i},i\in\mathbb{N}_{n}$. If this is true, then we say that $X_{i}$ is $p_{i}$ bounded by the pair $(a_{i},b_{i}),i\in\mathbb{N}_{n}$.

{\bf{Assumption 2 ($(p_{ij},k)$ hierarchy-bounded)}} Let $X_{i}$ be the independent random variable on a probability $(\Omega_{i},\mathcal{A}_{i},{\rm{P}}_{i}),$ $i\in\mathbb{N}_{n}$, s.t. there exists an integer $ k>1$, we have ${\rm{P}}_{i}(a_{ij}\leq X_{i}(\omega)\leq b_{ij})=p_{ij},j\in\mathbb{N}_{k},i\in\mathbb{N}_{n}$ and ${\rm{P}}_{i}(\bigcup_{j=1}^{k}(a_{ij}\leq X_{i}(\omega)\leq b_{ij}))=1,i\in\mathbb{N}_{n}$.  If this is true, then $X_{i}$ is $(p_{ij},k)$ hierarchy-bounded by the pair $(a_{ij},b_{ij}),j\in\mathbb{N}_{k},i\in\mathbb{N}_{n}$.

{\bf{Assumption 3 ($p$ difference-bounded)}}{\footnote{This assumption is similar to the assumption in \cite{Combes}.}} Let $g:\prod_{i=1}^{n}\Omega_{i}$ $\rightarrow\mathbb{R}$ be a function, s.t. for any $\ell\in\mathbb{N}_{n}$, $\exists A\subset\prod_{i=1}^{n}\Omega_{i}$, for any $\omega\in A$ and $\omega'\in A$ differ only in the $\ell$th coordinate, we have $|g(\omega)-g(\omega')|\leq c_{\ell}$ and ${\rm{P}}(A)=p$. If this is true, then $g$ is $p$ difference-bounded by $\{c_{\ell},\ell\in\mathbb{N}_{n}\}$.

{\bf{Assumption 4 ($(p_{j},k)$ hierarchy-difference-bounded)}} Let $g:\prod_{i=1}^{n}\Omega_{i}\rightarrow\mathbb{R}$ be a function, s.t. for any $\ell\in\mathbb{N}_{n}$,  there exists an integer $k>1$, $A_{j}\subset\prod_{i=1}^{n}\Omega_{i},j\in\mathbb{N}_{k}$, $\bigcup_{j=1}^{k}A_{j}=\prod_{i=1}^{n}\Omega_{i}$ and $A_{i}\cap A_{j}=\emptyset, i\not =j, i,j\in\mathbb{N}_{n}$, for any $\omega\in A_{j}$ and $\omega'\in A_{j}$ differ only in the $\ell$th coordinate, we have $|g(\omega)-g(\omega')|\leq c_{\ell j}$ and ${\rm{P}}(A_{j})=p_{j}$. If this is true, then $g$ is $(p_{j},k)$ hierarchy-difference-bounded by $\{c_{\ell j},\ell\in\mathbb{N}_{n}\},j\in\mathbb{N}_{k}$.

Under assumption 1 and 3, we study if Hoeffding's and McDiarmid's inequalities  still hold. Under Assumptions 2 and 4, we investigate if the convergence bounds of Hoeffding's and McDiarmid's inequalities can be better.

Actually, it is not difficult to see from Example 2 that it does not  always concentrate around its expectation.  For instance, ${\rm{P}}(f(X)=1)={1}/{(M^{n})}$ will be close to $0$ as $n$ tends to infinity. Therefore, Hoeffding's and McDiarmid's inequalities for such functions in Example 2 do not hold. Based on these assumptions, we will present several similar Hoeffding's and McDiarmid's inequalities (See Theorems 1, 2 and Corollaries 1, 2 in Section~5), which can  deal with unbounded and hierarchy-bounded functions.

At the end of this section, we compare four proposed bounded conditions with the previous three definitions in Section~2. Since the sums of random variables can be regarded as a special case of a multivariate random function, we will only discuss the relationships between the $p$ difference-bounded, $(p_{j},k)$ hierarchy-difference-bounded, uniformly differences-bounded, strongly differences-bounded and weakly differences-bounded conditions. Here we have the following pairwise comparisons $
(p_{j},k)\Longrightarrow{\rm{uniformly}}\Longrightarrow{\rm{strongly}}\Longrightarrow{\rm{weakly}}\Longrightarrow p$. Here the symbol ``$\Longrightarrow$'' means that the item is strictly stronger on the left than on the right. Therefore, from the above formal relation, it tells that the $p$ difference-bounded condition is the weakest and the $(p_{j},k)$ hierarchy-difference-bounded condition is the strongest.


\section{Extensions of Hoeffding's and McDiarmid's Inequalities}
In this section, we will show several extensions to Hoeffding's and McDiarmid's inequalities. 

Essentially, Hoeffding's inequality ({\ref{hoe}}) in Section~3 can be proved by combining the properties of convex functions, Taylor expansion, the monotonicity of probability measures, the exponential Markov inequality and the independence of random variables. Meanwhile, McDiarmid's inequality ({\ref{mcd}}) in Section~3 can be proved by constructing the martingale difference sequences in combination with a similar proof of Hoeffding's inequality. To extend these two concentration inequalities, we prove Theorems 1 and 2 using conditional mathematical expectation.  We assume that $X_{i}$ is independent random variable on a probability space $(\Omega_{i},\mathcal{A}_{i},{\rm{P}}_{i}), i\in\mathbb{N}_{n}$, and give the following condition:
%

{\bf{Condition 1 (Partition of product space)}} Let $\cup_{j=1}^{k}A_{ij}=\Omega_{i}, A_{ij'}\cap A_{ij''}=\emptyset, j'\not=j'', j'', j', j\in\mathbb{N}_{k},i\in\mathbb{N}_{n}$. Set{\small
\begin{equation}
\Psi_{n,k}=\{\prod_{i=1}^{n}A_{ij}| A_{ij}\in\{A_{i1},A_{i2},\ldots,A_{ik}\}\}
\end{equation}}the set $\Psi_{n,k}$ is called a partition of $\prod_{i=1}^{n}\Omega_i$.

We first provide two lemmas which will be used later.

{\bf{Lemma 1}}  Assume that $X_{i}$ is $({\rm{P}}_{i}(A_{ij}),k)$ hierarchy-bounded by the pair $(a_{ij},b_{ij})$, and Condition 1 holds. Let $\psi_{n,k}=\{(j_{1},j_{2},\ldots,j_{n})| j_{r}\in\mathbb{N}_{k}, r=1,2,\ldots,n\}$. Set $S_{n}=\sum_{i=1}^{n}X_{i}$ and $\widetilde{A}_{\bf j}=\prod_{i=1}^{n}A_{ij}\in\Psi_{n,k}$, $j\in\mathbb{N}_{k}$, ${\bf j}\in\psi_{n,k}$. Then, for any $t>0$, we have{\small
\begin{equation}{\label{aaaiaad1}}
\displaystyle {\rm{P}}(\{S_{n}-\sum\limits_{\bf j\in\psi_{n,k}}{\rm{E}}(S_{n}|\widetilde{A}_{\bf j}) I_{\widetilde{A}_{\bf j}}\geq t\}\cap \widetilde{A}_{{\bf j_{0}}})\leq {\rm{P}}(\widetilde{A}_{\bf j_{0}})\cdot\Delta_{3}
\end{equation}}where, $\Delta_{3}={\rm{exp}}({{-2 t^2}/{(\sum_{i=1}^{n}(a_{i{\bf j_{0}}}-b_{i{\bf j_{0}}})^{2}}})$, ${\bf j_{0}}=(j_{10},j_{20},\ldots,j_{n0})\in\psi_{n,k}$ is a constant vector, and we agreed that $a_{i{\bf j_{0}}}=a_{ij_{i0}}$ and $b_{i{\bf j_{0}}}=b_{ij_{i0}}$ $i\in\mathbb{N}_{n}$.

{\bf{Proof}} 
By the assumptions and definition of conditional mathematical expectation and the additivity and monotonicity of the probability measure, for any $t\geq 0$ and $s>0$, there exists a constant vector ${\bf j_{0}}$, we have{\small
\begin{equation}{\label{hold3}}
\displaystyle 
{\rm{P}}(\{S_{n}-\sum\limits_{\bf j\in\psi_{n,k}}{\rm{E}}(S_{n}|\widetilde{A}_{\bf j}) I_{\widetilde{A}_{\bf j}}\geq t\}\cap \widetilde{A}_{{\bf j_{0}}})\leq{\rm{exp}}\left({-st}\right)\cdot\Delta_{4}
\end{equation}}

Here, $\Delta_{4}={\rm{P}}(\widetilde{A}_{\bf j_{0}})\cdot \prod_{i=1}^{n}{\rm{exp}}({{s^2(b_{i{\bf j_{0}}}-a_{i{\bf j_{0}}})^2}/{8}})$. Take $s={4t}/{\sum_{i=1}^{n}(b_{i{\bf j_{0}}}-a_{i{\bf j_{0}}})^2}$, the inequality (\ref{hold3}) gets the minimum value ${\rm{P}}(\widetilde{A}_{{\bf j_{0}}})\cdot{\rm{exp}}\cdot\left({{-2 t^2}/{\sum_{i=1}^{n}(a_{i{\bf j_{0}}}-b_{i{\bf j_{0}}})^{2}}}\right)$. Finally, we have{\small
\begin{equation}{\label{final4}}
\displaystyle {\rm{P}}(\{S_{n}-\sum\limits_{\bf j\in\psi_{n,k}}{\rm{E}}(S_{n}|\widetilde{A}_{\bf j}) I_{\widetilde{A}_{\bf j}}\geq t\}\cap \widetilde{A}_{{\bf j_{0}}})\leq  {\rm{P}}(\widetilde{A}_{{\bf j_{0}}})\cdot\Delta_{3}
\end{equation}}

The proof of Lemma 1 is completed. 

Similarly, we have the following Lemma 2.

{\bf{Lemma 2}} Let the function $f$ be a map from $\mathcal{X}^{n}$ to $\mathbb{R}$. Assume that $f$ is ${\rm{P}}(A)$ difference-bounded by $\{c_{m},m\in\mathbb{N}_{n}\}$. Set $A=\prod_{i=1}^{n}A_{i}$. Then, for any $t>0$, we have{\small
\begin{equation}{\label{aaaiaad2}}
\begin{array}{ll}
&\displaystyle {\rm{P}}(\{f(X_{1},\ldots,X_{i},\ldots,X_{n})-{\rm{E}}(f(X_{1},\ldots,X_{i},\ldots,X_{n})\\
&\quad |A)\geq t \}\cap A)\\
\leq & \displaystyle {\rm{P}}(A)\cdot\Delta_{5}
\end{array}
\end{equation}}where $\Delta_{5}={\rm{exp}}({-2 t^2}/{(\sum_{i=1}^{n}c_{i}^{2})})$.

Denote $\mathcal{B}$ as the $\sigma-$algebra $\sigma(\Psi_{n,k})$, and $\mathcal{B}\subset\prod_{i=1}^{n}\mathcal{A}_i$.

{\bf{Theorem 1}} Under the assumptions of Lemma 1,  for any $t>0$, we have{\small
\begin{equation}{\label{exten3}}
\displaystyle {\rm{P}}(|S_{n}-{\rm{E}}(S_{n}|\mathcal{B})|\geq t)\leq 2\cdot
\sum\limits_{\bf j\in\psi_{n,k}}{\rm{P}}(\widetilde{A}_{\bf j})\cdot\Delta_{6}
\end{equation}}where $\Delta_{6}={\rm{exp}}({{-2 t^2}/({\sum_{i=1}^{n}(a_{i{\bf j}}-b_{i{\bf j}})^{2}})})$, and we agreed that $a_{i{\bf j}}=a_{ij_{i}}$ and $b_{i{\bf j}}=b_{ij_{i}}$, $i\in\mathbb{N}_{n}$.

{\bf{Proof}}
From the assumptions, we have{\small
\begin{equation}{\label{addijca1}}
\displaystyle\bigcup\limits_{B\in\Psi_{n,k}} B=\bigcup\limits_{\bf j\in\psi_{n,k}} \widetilde{A}_{\bf j}  =\prod_{i=1}^{n}\Omega_i
\end{equation}}

By the equation (\ref{addijca1}) and definition of conditional mathematical expectation and the additivity and monotonicity of the probability measure,  for any $t\geq 0$, we have{\small
\begin{equation}
{\rm{P}}(S_{n}-{\rm{E}}(S_{n}|\mathcal{B})\geq t)=\sum\limits_{\bf j'\in\psi_{n,k}}\Delta_{7}
\end{equation}}where$\Delta_{7}={\rm{P}}(\{S_{n}-\sum_{\bf j\in\psi_{n,k}}{\rm{E}}(S_{n}|\widetilde{A}_{\bf j}) I_{\widetilde{A}_{\bf j}}\geq t\}\cap \widetilde{A}_{\bf j'})$.

By Lemma 1, it is easy to show that Theorem 1 holds.

{\bf{Theorem 2}} Let the function $f$ be a map from $\mathcal{X}^{n}$ to $\mathbb{R}$. Assume that $f$ is $({\rm{P}}_{i}(A_{ij}),k)$ hierarchy-difference-bounded by $\{c_{mj},m\in\mathbb{N}_{n}\},j\in\mathbb{N}_{k}$, and Condition 1 holds. Let $\psi_{n,k}=\{(j_{1},j_{2},\ldots,j_{n})| j_{r}\in\mathbb{N}_{k}, r=1,2,\ldots,n\}$. Set $\widetilde{A}_{\bf j}=\prod_{i=1}^{n}A_{ij}\in\Psi_{n,k}$, $j\in\mathbb{N}_{k}$, ${\bf j}\in\psi_{n,k}$. For any $t>0$, we have{\small
\begin{equation}{\label{exten7}}
\displaystyle {\rm{P}}(|f(X_{1},\ldots,X_{n})-{\rm{E}}(f(X_{1},\ldots,X_{n})|\mathcal{B})|\geq t)\leq 2\cdot\Delta_{8}
\end{equation}}where $\Delta_{8}=\sum_{\bf j\in\psi_{n,k}}{\rm{P}}(\widetilde{A}_{\bf j})\cdot{\rm{exp}}({{-2 t^2}/({\sum_{i=1}^{n}c_{m{\bf j}}^{2}})})$, and we agreed that $c_{i{\bf j}}=c_{mj_{m}}, m\in\mathbb{N}_{n}$.

{\bf{Proof}} By combining the Lemma 2 and the methods employed in Theorem 1, Theorem 2 can be easily proved.

From the above Theorems, we have the following corollaries:

{\bf{Corollary 1}} Assume that $X_{i}$ is ${\rm{P}}_{i}(A_{i})$ bounded by the pair $(a_{i},b_{i}),i\in\mathbb{N}_{n}$. Set $S_{n}=\sum_{i=1}^{n}X_{i}$ and $A=\prod_{i=1}^{n}A_{i}$. Then, for any $t>0$, we have {\small
\begin{equation}{\label{exten1}}
\displaystyle {\rm{P}}(|S_{n}-{\rm{E}}(S_{n}|A)|\geq t)\leq\displaystyle 2\cdot{\rm{P}}(A)\cdot\Delta_{9}+1-{\rm{P}}(A)
\end{equation}}where $\Delta_{9}={\rm{exp}}({{-2 t^2}/{(\sum_{i=1}^{n}(a_{i}-b_{i})^{2})}})$.

{\bf{Corollary 2}} Under the assumptions of Lemma 2. Then, for any $t>0$, we have {\small
\begin{equation}{\label{exten5}}
\displaystyle {\rm{P}}(|f(X_{1},\ldots,X_{n})-{\rm{E}}(f(X_{1},\ldots,X_{n})|A)|\geq t)\\
\displaystyle\leq 2\cdot\Delta_{10}+1-{\rm{P}}(A)
\end{equation}}where $\Delta_{10}={\rm{P}}(A)\cdot {\rm{exp}}({{-2 t^2}/({\sum_{i=1}^{n}c_{i}^{2}})})$.

{\bf{Note 4}} From the above theorems and corollaries, we can conclude that:

$\blacksquare$ Under the four assumptions proposed by us, the random variable or multivariate random function does not concentrate around its mathematical expectation. Theorems 1 and 2 imply that the random variable or multivariate random function should concentrate around its conditional expectation. And to some extent, Corollaries 1 and 2 also imply such concentration in these two cases: 1) $1-{\rm{P}}(A)$ is close to $0$ as $n$ tends to $\infty$, see Example 3, or 2) $A=\Omega$.

$\blacksquare$ The original inequalities (\ref{hoe}) and (\ref{mcd}) can be viewed as special cases of the extensions (\ref{exten1}) and (\ref{exten5}): if $A$ increases up to $\Omega$, then the inequality (\ref{exten1}) reduces to the inequality (\ref{hoe}) in Section~3. If $A$ increases up to $\Omega$, then the inequality (\ref{exten5}) reduces to the inequality (\ref{mcd}) in Section~3. Here $A$ equals to $\prod_{i=1}^{n}A_{i}$, and $\Omega$ equals to $\prod_{i=1}^{n}\Omega_{i}$. In addition, the bounds of the extensions (the equalities (\ref{exten1}) and (\ref{exten5})) improve the bounds of the equation (\ref{comb}) given by \cite{Combes}: 1) the bound of the equation (\ref{comb}) is trivial if the item $p$ in the equation (\ref{comb}) is larger than $1/2$, whereas ${\rm P}(A)$ in our extensions has no such limitations, 2) the factor of the item `$\rm{exp}$' in the equation (\ref{comb}) is always $1$, whereas our factor is a probability ${\rm P}(A)$.

$\blacksquare$ The bounds of the extensions (the equalities (\ref{exten3}) and (\ref{exten7})) are tighter than the equalities (\ref{hoe}) and (\ref{mcd}): let $a'_{i{\bf j}}=a_{i{j_{i}}}=\min_{j\in\{j_{1},\ldots,j_{n}\}}a_{ij}$
and
$b'_{i{\bf j}}=b_{i{j_{i}}}=\max_{j\in\{j_{1},\ldots,j_{n}\}}b_{ij},i\in\mathbb{N}_{n}$ where $a_{ij_{i}}\leq b_{ij_{i}},i\in\mathbb{N}_{n}$. Then we have
{\small
\begin{equation}{\label{modifyadd1}}
\displaystyle\sum\limits_{{\bf j}\in\psi_{n,k}}P(\widetilde{A}_j)\cdot {\rm{exp}}\left({{-2 t^2}/{\sum\limits_{i=1}^{n}(a_{i{\bf j}}-b_{i{\bf j}})^{2}}}\right)\\
\displaystyle\leq\Delta_{11}
\end{equation}}where $\Delta_{11}={\rm{exp}}({{-2 t^2}/{\sum_{i=1}^{n}(a'_{i{\bf j}}-b'_{i{\bf j}})^{2}}})$.

According to definition of conditional expectation, $\mathcal{B}=\{\emptyset,\Omega\}$ implies ${\rm{E}}(X|\mathcal{B})={\rm{E}}(X)$. By substituting the inequality (\ref{modifyadd1}) into the inequality (\ref{exten3}), we thus get{\small
\begin{equation}{\label{add1}}
\displaystyle {\rm{P}}(|S_{n}-{\rm{E}}(S_{n})|\geq t)\\
\displaystyle\leq\displaystyle 2\cdot{\rm{exp}}\left({{-2 t^2}/{(\sum\limits_{i=1}^{n}(a'_{i{\bf j}}-b'_{i{\bf j}})^{2})}}\right)
\end{equation}}
The inequality (\ref{add1}) means that, if we take $a'_{i{\bf j}}=a_{i{j_{i}}}$ and $b'_{i{\bf j}}=b_{i{j_{i}}}$, $i\in\mathbb{N}_{n}$ in Theorem 1, then the inequality (\ref{exten3}) will reduce to the inequality (\ref{hoe}) in Section~3. Thus, we conclude that if the bound of the random variables is refined on a sample space $\Omega$, then a tighter bound by Theorem 1 will be obtained.

Similarly, if we take $c'_{m{\bf j}_{m}}=c_{m{j_{m}}}=\max_{j\in\{j_{1},\ldots,j_{n}\}}b_{mj}$, $m\in\mathbb{N}_{n}$ in Theorem 2, then the inequality (\ref{exten7}) will reduce to the inequality (\ref{mcd}) in Section~3, leading to the conclusion that if the bounded differences of the function $f$ is refined on a sample space $\Omega$, then a tighter bound by Theorem 2 will be reached.

{\bf{Note 5}} For unbounded random variables, there are also some Bernstein-like results \cite{Geer,Gine}. These results all require that the moment (for example, variance) exists or is uniformly bounded, and this limits their extension to some applications, whereas our results have no restrictions on the moment based on Corollary 1 or 2.

\section{Applications in Statistical Learning Theory}
In the previous section, we have proposed several extensions and compared them with existing bounds. Now we discuss these extensions to applications in learning theory through four examples. We show that our extensions are slightly faster than the existing results in some special cases and artificially bounding the unbounded loss function may not discover the overfit. 

{\bf{Example 3}} {\cite{Combes}} gave an example as follows:

Let $\Omega=\{0,1\}^{n}$, $X_{i}$ follows a Bernoulli distribution $Bern(1,p)$, $i=1,2,\ldots, n$, $A=\Omega\backslash\{(0,\ldots, 0),(1,\ldots,1)\}$, there exists a constant $B\geq 0$. Set $f$ a piecewise function: if $X_{i}=0, i=1,2,\ldots,n$, $f(X_{1},\ldots,X_{n})=B$; if $X_{i}=1, i=1,2,\ldots,n$, $f(X_{1},\ldots,X_{n})=-B$; otherwise, $f(X_{1},\ldots,X_{n})=({1}/{n})\sum_{i=1}^{n}2(X_{i}-1)$.
Then, {\cite{Combes}} obtained the generalization bound as follows{\small
\begin{equation}
{\rm P}(f(X)\geq t)\leq 2^{-n}+{\rm exp}\left(-\frac{n}{2}(t-2^{1-n})^{+2}\right)
\end{equation}}
From Corollary 1, we have the following generalization bound{\small
\begin{equation}
{\rm P}(f(X)\geq t)\leq 2^{-n}+(1-2^{-n})\cdot{\rm exp}\left(-\frac{n}{2}t^2\right)
\end{equation}}
\begin{figure}
\centering{
\begin{minipage}[b]{1.5\textwidth}
\includegraphics[height=0.15\textwidth,width=.3\textwidth]{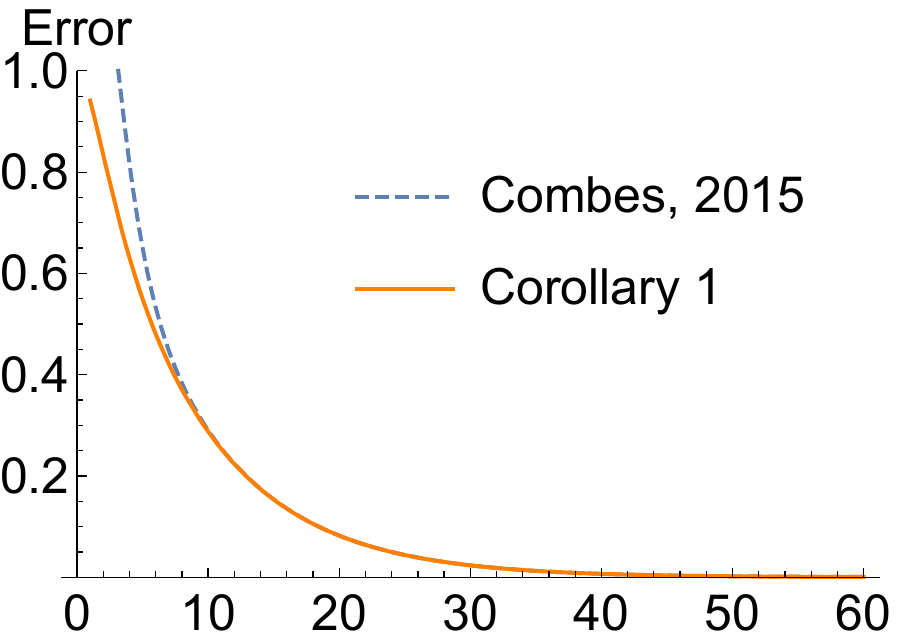}\qquad\includegraphics[height=0.15\textwidth,width=.3\textwidth]{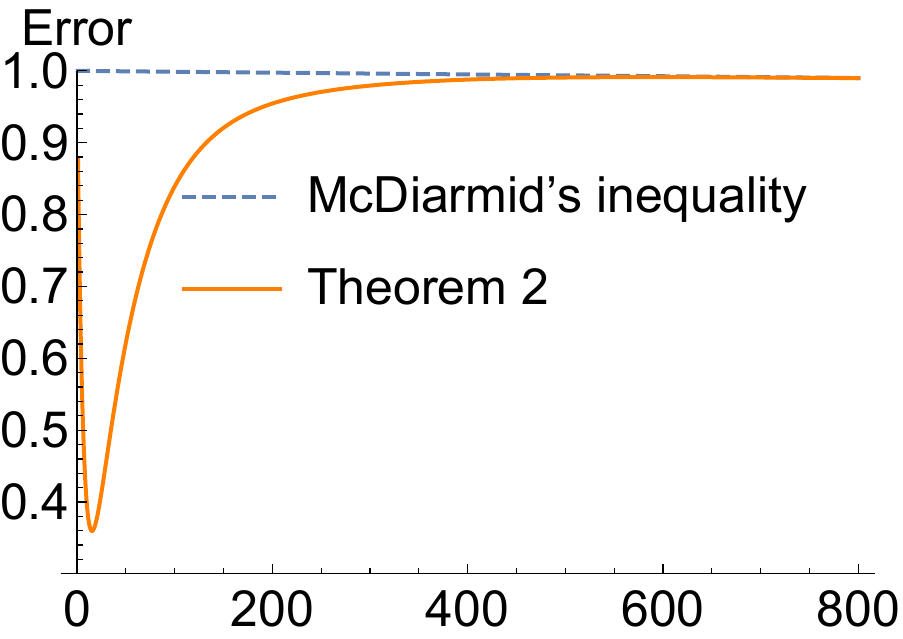}
\end{minipage}
\\[0.1cm]
\begin{minipage}[b]{1.5\textwidth}
\includegraphics[height=0.15\textwidth,width=.3\textwidth]{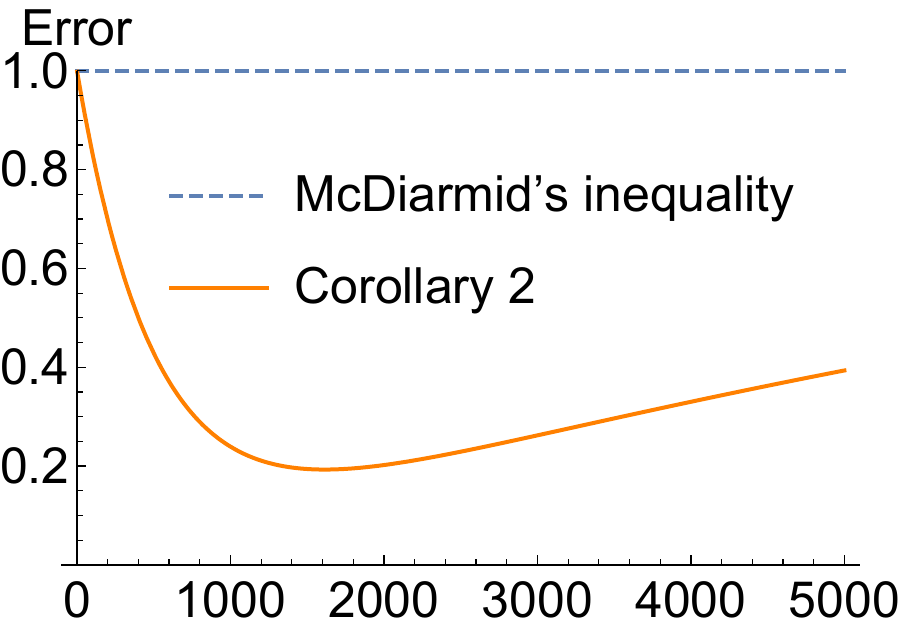}\qquad
\includegraphics[height=0.15\textwidth,width=.3\textwidth]{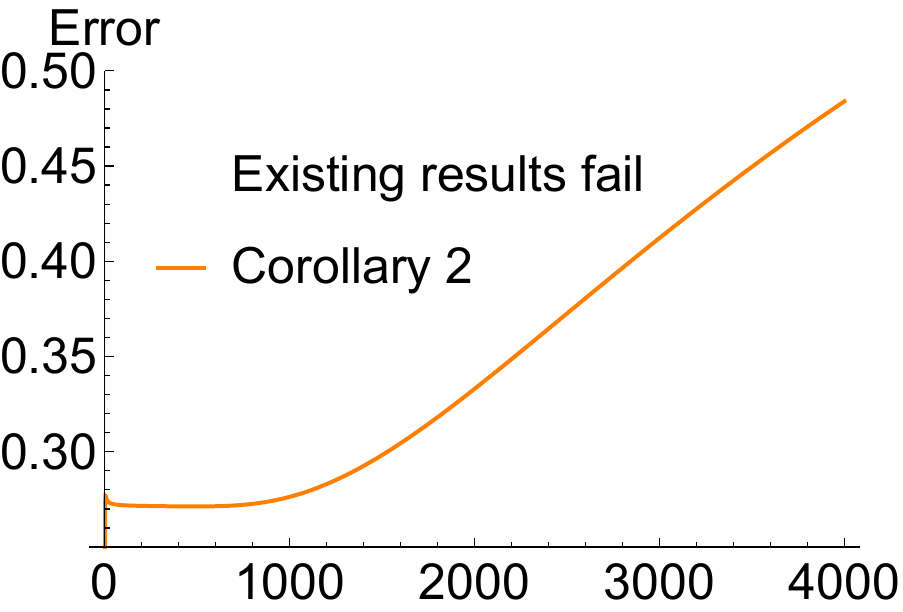}
\end{minipage}}
\caption{Here the horizontal axis means the sample complexity. From left to right, 1) The error probabilities from Corollary 1 and [Combes, 2015] (Take $t=0.5$). 
2) The error probability from Theorem 2 is slightly faster than the error probability from McDiarmid's inequality (Take $t=25$). 3) The error probability from Corollary 2 is slightly faster than that from McDiarmid's inequality (Take $t=3$). 4) The relation between the sample complexity and the error probability from Corollary 2, and the existing results fail (Take $\phi(n)=n^{1001/1000}$ and $t=50$).}
\end{figure}
From Figure 1.(1) it can be seen that the error probability from Corollary 1 is slightly faster than the error probability from \cite{Combes}.

{\bf{Example 4}} We assume that $\Omega=\{0,1,\ldots,97,1000,10000\}^{n}$, $X_{i}$ follows a multinomial distribution $Mult(100,{\bf p}), {\bf p}=(1/100,1/100,\ldots,1/100)$, $i=1,2,\ldots, n$. Let $f(X_{1},\ldots,X_{n})=\sum_{i=1}^{n}X_{i}$. By the assumptions, we have ${\rm P}(|f(X_{1},\ldots,X_{i},\ldots,X_{n})-f(X_{1},\ldots,X'_{i},\ldots,X_{n})|\leq 97/n) =(1-2/100)^{n}$,  ${\rm P}(97/n\geq |f(X_{1},\ldots,X_{i},\ldots,X_{n})-f(X_{1},\ldots,X'_{i},\ldots,X_{n})|\leq 1000/n)=(1-1/100)^{n}-(1-2/100)^{n}$ and ${\rm P}(1000/n\geq |f(X_{1},\ldots,X_{i},\ldots,X_{n})-f(X_{1},\ldots,X'_{i},\ldots,$ $X_{n})|\leq 10000/n)=1-(1-1/100)^{n}$. Then, from the equation ($\ref{mcd}$), we have{\small
\begin{equation}
\displaystyle {\rm{P}}(f(X_{1},\ldots,X_{n})-{\rm{E}}(f(X_{1},\ldots,X_{n})|\mathcal{B})\geq t)\leq \Delta_{12}
\end{equation}}where $\Delta_{12}={\rm{exp}}({{-2 t^2}/({\sum_{i=1}^{n}(10000/n)^{2}})})$.

From Theorem 2 we have{\small
\begin{equation}
\displaystyle {\rm{P}}(f(X_{1},\ldots,X_{n})-{\rm{E}}(f(X_{1},\ldots,X_{n})|\mathcal{B})\geq t)\leq \Delta_{13}+\Delta_{14}+\Delta_{15}
\end{equation}}where $\Delta_{13}=(1-2/100)^{n}\cdot{\rm{exp}}({{-2 t^2}/({\sum_{i=1}^{n}(97/n)^{2}})})$, $\Delta_{14}=((1-1/100)^{n}-(1-2/100)^{n})\cdot{\rm{exp}}({{-2 t^2}/({\sum_{i=1}^{n}(1000/n)^{2}})})$ and $\Delta_{15}=(1-(1-1/100)^{n})\cdot{\rm{exp}}({{-2 t^2}/({\sum_{i=1}^{n}(10000/n)^{2}})})$. The values of $(1-2/100)^{n}$, $((1-1/100)^{n}-(1-2/100)^{n})$ and $(1-(1-1/100)^{n})$ can be seen as the weights obtained according to the proportion of samples.
%

From Figure 1.(2) it can be seen that the error probability from Theorem 2 is firstly faster than that from McDiarmid's inequality, and then the distinction of their error probabilities gradually becomes less visible.

{\bf{Example 5}} We assume that $\Omega=\{0,1,\ldots,97,98,\infty\}^{n}$, $X_{i}$ follows a multinomial distribution $Mult(100,{\bf p}), {\bf p}=(101/10000,\ldots,101/10000,1/10000)$, $i=1,2,\ldots, n$. Let $f(X_{1},\ldots,X_{n})=\sum_{i=1}^{n}X_{i}$. By the assumptions, we have ${\rm P}(|f(X_{1},\ldots,X_{i},\ldots,X_{n})-f(X_{1},\ldots,X'_{i},\ldots,X_{n})|\leq 98/n) =(1-1/10000)^{n}$. Then, from Corollary 2 we have{\small
\begin{equation}
\displaystyle {\rm{P}}(f(X_{1},\ldots,X_{n})-{\rm{E}}(f(X_{1},\ldots,X_{n})|\mathcal{B})\geq t)\leq \Delta_{16}
\end{equation}}where $\Delta_{16}=(1-1/10000)^{n}\cdot{\rm{exp}}({{-2 t^2}/({\sum_{i=1}^{n}(98/n)^{2}})})+(1-1/10000)^{n}$.

From Figure 1.(3) it can be seen that the error probability decreases as the sample complexity increases when the sample complexity is smaller than $1500$. When the sample complexity is larger than $1500$, meanwhile, the error probability increases with the increase of the sample complexity. However, the result from McDiarmid's inequality is trivial. Examples 1 and 2 in Section 4 can be analyzed similarly.

For simplicity, in the discussions of Examples 3-5 we do not involve loss functions. In the last example, we will introduce a loss function and show how to apply our extensions.

{\bf{Example 6}} We assume that the linear model for regression is $y=h(x)+\epsilon$, where $\epsilon$ is a standard Cauchy random variable with the density function $({1}/{\pi})\cdot{1}/({1+x^{2}})$. The loss function $L$ is defined by the absolute loss $|h(x)-y|, h\in\mathcal{H}$ (Denoted by $Q(h,z)$).

It is obvious that the expected value of $\epsilon$ does not exist. Therefore, Hoeffding's inequality and McDiarmid's  inequality do not hold because Hoeffding's inequality and McDiarmid's inequality are distribution independent. Here, our results will be valid. We can employ Corollary 2 to analyze its generalization bound.

Let the set $A_{i}$ in Corollary 2  be  ${-\phi(n)\leq\epsilon_{i}\leq \phi(n)}, i=1,2,\ldots,n$. Then, we have  
{\small
\begin{equation}{\label{AAMASadd3}}
\displaystyle {\rm{P}}(\frac{1}{n}\cdot\sum_{i=1}^{n}Q(h,z_i)-{\rm{E}}(Q(h,z)|A)\geq t)\leq\Delta_{17}
\end{equation}}where $\Delta_{17}=\Delta_{18}^{n}\cdot{\exp{(-{2\cdot n\cdot t^2}/{\phi^{2}(n)})}}+1-\Delta_{18}^{n}$, $\Delta_{18}=({1}/{2} +{\arctan(\phi(n))}/{\pi})$.

From Figure 1.(4) it can be seen that the error probability is smaller than $0.5$ when the sample complexity is smaller than $4500$. When the sample complexity is smaller than $4500$, the error probability becomes larger than $0.5$, and tends to $1$ with the increase of the sample complexity.
 
%
{\bf{Note 6}} Examples 3 and 4 indicate that our extensions are slightly faster than the existing results when the sample complexity is not so large. Examples 5 and 6
show that our extensions describe that the error probability evolves over the sample complexity while the existing results are failure or trivial. Furthermore, from Example 4, we find that the effect of weighting on the proportion of samples is effective when the sample complexity is small, and becomes negligible when the complexity of the sample increases. From examples 5 and 6, we can tell that the generalization analysis on the artificial bound of the unbounded loss function (or the loss function whose expectation does not exist) may not catch the overfit, and for the generalization analysis of these loss functions. However, the classical learning framework may not be enough for the generalization analysis of these loss function. Such a phenomenon has also been observed by~\cite{MendelsonL}.

\section{Conclusion}\label{sec7}
In this paper, we review the conditions and limitations of Hoeffding's and McDiarmid's inequalities. We propose four new conditions and compare them with the existing difference-bounded conditions. Based on our proposed conditions, we obtain several extensions of Hoeffding's and McDiarmid's inequalities. Through four examples, we also discuss the potential applications of our extensional results in learning theory. As future work, we will study how to design effective machine learning algorithms to analyze the problems in learning theory based on the proposed extensions.

{\small{
\bibliographystyle{unsrt}
\bibliography{\jobname}
}}
\end{document}